\documentclass{article}

\usepackage[table,xcdraw]{xcolor}

\usepackage{listings}
\renewcommand{\figurename}{Fig.}

\usepackage{amsmath}

\usepackage{authblk}

\usepackage{PRIMEarxiv}

\usepackage[utf8]{inputenc} 
\usepackage[T1]{fontenc}    
\usepackage[hidelinks]{hyperref}       
\usepackage{url}            
\usepackage{booktabs}       
\usepackage{amsfonts}       
\usepackage{nicefrac}       
\usepackage{microtype}      
\usepackage{lipsum}
\usepackage{fancyhdr}       
\usepackage{graphicx}       
\graphicspath{{media/}}     

\title{Content Rating Classification for Fan Fiction
}

\author[1]{Yu Qiao (\href{https://orcid.org/0000-0003-4738-9642}{0000-0003-4738-9642}) }
\author[1]{James Pope (\href{https://orcid.org/0000-0003-2656-363X}{0000-0003-2656-363X}) }

\affil[1]{University of Bristol, Bristol, UK}

\begin{document}
\maketitle

\begin{abstract}

Content ratings can enable audiences to determine the suitability of various media products.  With the recent advent of fan fiction, the critical issue of fan fiction content ratings has emerged.  Whether fan fiction content ratings are done voluntarily or required by regulation, there is the need to automate the content rating classification.  The problem is to take fan fiction text and determine the appropriate content rating.  Methods for other domains, such as online books, have been attempted though none have been applied to fan fiction.  We propose natural language processing techniques, including traditional and deep learning methods, to automatically determine the content rating.  We show that these methods produce poor accuracy results for multi-classification.  We then demonstrate that treating the problem as a binary classification problem produces better accuracy.   Finally, we believe and provide some evidence that the current approach of self-annotating has led to incorrect labels limiting classification results.

\end{abstract}


\section{Introduction}
\label{sec:introduction}

Content ratings are widely used to determine which age group is suitable to watch media content. The film industry is widely known to use such classifications for movies. Before that time, there had been censorship about films or comics but aimed at filtering unsuitable works, not rating or giving suggestions. In 1968, the Motion Picture Association(MPA) created their rating system in the United States and its territories to replace the Motion Picture Production Code \cite{williams2007prevalence}, which defines four categories:

\begin{itemize}
    \item Rated G: General audiences can watch this film.
    \item Rated M: Mature audiences are suggested to watch this film. Children should be accompanied by parents to watch.
    \item Rated R: Restricted – Persons under 16 are not admitted to watching this film unless a parent or adult guardian accompanies them.
    \item Rated X: Persons under 16 are not permitted to watch this film.
\end{itemize}

Although the definition and count of the categories have been modified several times, the rating covers the age from a child to an adult. Also, the rating components include violence, language, substances, nudity and sex. These components are inherited by content rating systems in other fields and followed by film, comics, television, and video games. Music and the Internet applied the rating.



Fan fiction is the crossing field of literature and the Internet.  Fans of a work of fiction write their own stores related to the original theme. Although fan fiction is a kind of media, consistent content ratings for fan fiction are missing. Each website may use a different rule and many fan fiction works do not have a content rating.  We consider the archiveofourown website that allows fan fiction authors to self-annotate the content rating.  Nevertheless, most of the works do not have a content rating.  The problem is to automate the content rating of fan fiction.

In this paper we first propose and evaluate a multi-classification approach with traditional and deep learning models using the self-annotations from archiveofourown.  The results are somewhat poor so we next consider addressing the problem using binary classification.  After model tuning, we compare the modelling approaches.  We show that accuracy typically improves with more text in each instance and determine that the traditional approaches can achieve an accuracy of nearly 70\%.  We find that a number of self-annotations are incorrect and believe that this accuracy can be improved with accurate labels.





The paper is structured as follows: Section \ref{sec:related_work} briefly describes the related work, section \ref{sec:dataset} describes the dataset, section \ref{sec:methods} details the various models, section \ref{sec:results} includes the results and analysis, and section \ref{sec:conclusion} concludes our work.

\section{Related Work}
\label{sec:related_work}
A typical natural language processing (NLP) pipeline involves transforming text into a vector and then using those vectors with various classifiers.  For converting text into a vector, Salton, et al. \cite{SaltonTFIDF1994}, make a notable extension to the bag of words (BOW) approach by adding weights based on term and document frequency (TF-IDF).  We consider TF-IDF along with the Naive Bayes (NB) and Random Forest (RF) \cite{breiman2001random} classifiers as traditional approaches.  Subsequently, \textit{word2vec} \cite{WORD2VEC2013}, describes  how words and their surrounding words can be transformed into a vector (a.k.a. word embedding).  Melamud, et al. \cite{CONTEXT2VEC} propose passing these word embeddings to a Bidirectional Long Short Term Memory (Bi-LSTM) neural network.  The Bidirectional Encoder Representations from Transformers (BERT) \cite{BERT} produce contextualised word embeddings.  ALBERT \cite{lan2019albert} uses  parameter reduction techniques to improve the computational performance of BERT and introduce a self-supervised loss that focuses on capturing inter-sentence coherence.  Similarly, DistilBERT \cite{sanh2019distilbert} leverages knowledge distillation during the pre-training phase resulting in a reduced model size.

Mahsa \cite{shafaei2020age} used dialogues to predict the MPAA rating. Sensitive word systems have been set to detect abusive or mature content in online media \cite{wanner2011my}.  Classifying the content rating of online books has previously been researched \cite{brewer2021identifying}. The model predicts a book’s content rating level through seven categories.  Shafaei, et al. \cite{shafaei2020age}, make the first attempt to use text classification to build a content rating system. They use movie scripts to predict the Motion Picture Association of America (MPAA) rating, a film rating system that establishes the appropriate age for movie viewers. They use 300-dimensional pre-trained Glove embedding to extract features and then use LSTM with attention to training the model.  Donaldson and Pope \cite{DONALD2022} propose using natural language processing techniques to distinguish fan from print fiction.

To our knowledge, our research is the first to apply these NLP techniques to fan fiction content rating classification.

\section{Dataset}
\label{sec:dataset}
\begin{figure}[h]
    \centering
    \includegraphics[width=\linewidth]{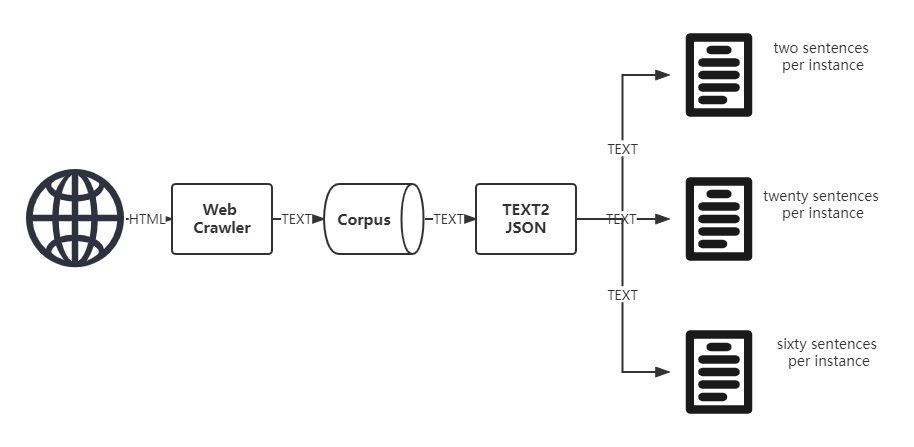}
    \caption{Data Collection Overview}
    \label{fig:datacollect}
\end{figure}

This phase collects data from the website, generates the sample set and stores it using JSON format. \figurename~\ref{fig:datacollect} depicts the data collection pipeline.

\subsection{Data Source}
The fan fiction corpus was collected from the online platform archiveofourown.org, which is a fan-created, fan-run, nonprofit, noncommercial archive for fan works. Until now, it owns over 9,744,000 works. The works may be motivated by books, movies, games etc. Since the creations are expressed in words, they are considered fan fiction. This website uses five kinds of labels to mark the works: General Audiences, Teen And Up Audiences, Mature, Explicit or Not Rated, which supports the following supervised machine learning.

\subsection{Data collection}
We modify the web crawler provided by Donaldson \cite{DONALD2022} to collect the data. The pipeline in the Figure \ref{fig:datacollect} illustrates the process. The first step is to crawl the data from archiveofourown. Fan fiction works on the website were downloaded as HTML files and then converted to text (UTF-8) format.  Works with the same rating were stored in the folder named with the rating to locate the class and build the sample set. To filter the unnecessary fiction, the crawler only accepted works whose language is English. After that, texts in each folder were generated and converted into JSON format for further processing. To test the influence of the sentence number in each sample, we created 60 separate JSON files whose filenames corresponded to the number of sentences in each instance. It began at two sentences per instance and increased by two sentences by a step. When it reached 20, the step became four sentences. Small text files which contained little information were filtered.  Listing \ref{lstexample} depicts an example instance with 10 sentences classified as \textit{G}.

\vspace{20pt}

\begin{lstlisting}[label=lstexample,caption=Example G Instance (10 sentences/instance),frame=tb]
I'm sorry it had to end this way.
I love you until the end of time.
She turns to face you, tears starting to flow down her gorgeous face.
You've seen her bruised and bloodied and at her wit's end and even all but broken. 
But you shouldn't see her like this.
She shouldn't be like this.
Why is she so calm, facing certain death that will end even an immortal?
You have no answer and can only wish her goodbye until you reach oblivion.
It was the only way to save everyone. Why, now, when the choice is already made, do you start to feel regret? 
Perhaps it is because you know that you will never see her again.
\end{lstlisting}

\subsection{Data Preparation}
In this step, the data preparation block balanced the sample size in each category and normalises the text. Samples were first grouped by category and counted. Next, we extracted half of the minimum sample size of the samples in each category. For example, if the minimum category owns 10,000 instances, we would randomly choose 5,000 instances from each category. The sample reduced indeterminacy by balancing the data and increasing randomness. The text was lowercased, stop words were removed (using Rainbow set), proper nouns and punctuation were removed, and finally the text was  lemmatised. A stemming operation may cause words to lose their meaning if it is in conflict with lemmatization. Therefore, it was not adopted. After the processing, samples were divided to train and test set on a 70/30 split for further use.

\section{Methods}
\label{sec:methods}

This section first introduces the classical and deep learning methods and their variants. The section concludes by presenting the F1 score evaluation metric.

\subsection{ Classical Methods }

\begin{figure}[ht]
    \centering
    \includegraphics[width=\linewidth]{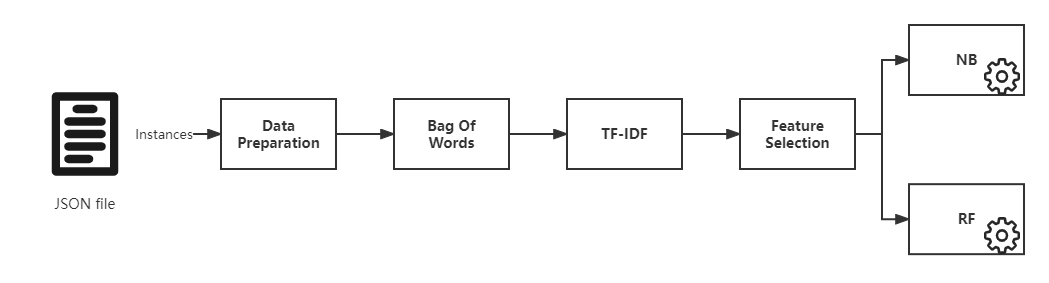}
    \caption{Classic Methods Pipeline}
    \label{fig:classicpipeline}
\end{figure}

Term frequency, inverse document frequency (TF–IDF) \cite{robertson2004understanding} is an effective document representation that applies weights to the bag-of-word scheme. This scheme assumes that the words used in a document can represent the document, which produces TF. The other assumption is that words that repeatedly appear in a document but rarely appear in other documents should be considered vital for that document. The IDF is associated with this assumption. The $TF_{ij}$ parameter is defined as the occurrence time of the word $i$ in document $j$; a larger value means a more important word. The $DF_i$ parameter is the number of documents in which the word $i$ appears at least once; a larger value indicates a more common word. If a word have a large TF and a small DF, it can be considered important for a document. On the other hand, articles like ‘a’ and ‘the’ and pronouns like ‘it,’ ‘this,’ ‘that,’ and ‘those’ frequently appear in each document but cannot be considered important because they are common in all documents. Hence, TF–IDF is defined by:

\begin{equation}
    \text{TF–IDF}_{i,j} = \text{TF}_{i,j} \times log(N/DF_i)
\end{equation}

We follow TF-IDF by feature selection.  The chi-squared ($\chi^2$) test is generally used to examine the independence of two events \cite{liu1995chi2}.  The chi-squared score is used to reduce features that are not statistically relevant to the label.  The p-value cutoff to choose which features to keep is a critical hyper-parameter.

After feature selection, two classifiers are considered.  The naive Bayes (NB) classifier \cite{rish2001empirical} is the canonical probabilistic classifier. The basic assumption of NB is that all features of the examples are independent of each other given the class. This assumption makes the naive Bayes method simple but sometimes sacrifices certain classification accuracy.  The random forest \cite{breiman2001random} consists of decision trees. It applies the general technique of bootstrap aggregating or bagging to tree learners. It will create several decision tree models.  Each tree will randomly choose samples from the sample set and some features from all features to learn.  The number of trees and tree depth are critical hyper-parameters.

\subsection{ Deep Learning Methods }

\subsubsection{ Word2Vec + LSTM }
This subsection introduces the Word2Vec + LSTM model.  This block first counts the word frequency, and n-gram phrases were also considered to be words. Next, it numbered each word in the input text according to the descending order of the word frequency. After that, it padded or truncated the words to fix the text length. Each sentence was allocated five words. For example, the text length would be ten if there were two sentences per instance. Only words whose frequency order was in the top 30000 were considered important; other words were filtered out. After the text was transferred to sequence, the pre-trained Word2Vec model "glove-wiki-gigaword-300" transformed it to become word embeddings. In the end, the deep learning model based on LSTM adopted the embeddings as input, trained the model and made the prediction.

\begin{table}
\centering
\caption{Word2Vec + LSTM Model}
\label{tab:word2vev-lstm}
\begin{tabular}{ll}
\hline
\rowcolor[HTML]{C0C0C0} 
Layer     & Parameters                            \\ \hline
Input     & textLength=5 x sentences/instance    \\  \hline
Embedding & output=(textLength, 300)             \\  \hline
Attention & output=(textLength, 300)             \\  \hline
LSTM      & output=(textLength, 2*textLength)  \\ 
          & dropout=0.2.                         \\  \hline
LSTM      & output=(2*textLength)              \\
          & dropout=0.2                          \\  \hline
Dense     & activation=ReLU, output=(64)         \\  \hline
Dense     & activation=Softmax, output=(2)       \\  \hline
\end{tabular}
\end{table}

\subsubsection{ BERT + Deep Learning Model }
Table \ref{tab:bert-dlm} introduces the deep learning model that includes a BERT \cite{devlin2018bert} layer to embed the input sentences.  The first layer truncates/pads the instances, masks, and builds inputs needed for the BERT layer. In the BERT layer, three kinds of the pre-trained model were tested, but the shape kept the same. Next, a layer was used to process the output of the BERT. Finally, the prediction was made after two dense layers.


\begin{table}
\centering
\caption{BERT + DLM Model}
\label{tab:bert-dlm}
\begin{tabular}{ll}
\hline
\rowcolor[HTML]{C0C0C0} 
Layer     & Parameters                         \\ \hline
Input     & textLength=5 x sentences/instance  \\
input idx  &                                   \\ \hline
Input     & textLength=5 x sentences/instance  \\
input masks  &                                 \\ \hline
BERT      & last hidden state=(textLength, 768) \\ \hline
LSTM      & output=(2 x textLength)            \\
Pooling   & output = 768                       \\ \hline
Dense     & activation=ReLU, output=(64)       \\ \hline
Input     & textLength=5 x sentences/instance  \\
input segments  &                              \\ \hline
Dense     & activation=Softmax, output=(2)     \\ \hline
\end{tabular}
\end{table}



\subsection{ Accuracy Metric }

In a binary classification task, instances are labelled as positive or negative, and the predictions are considered to be true or false. Given the actual and predicted class, there are four possibilities: true positive (TP), true negative (TN), false positive (FP), and false negative (FN).  Accuracy is calculated as follows:

\begin{equation}
    Accuracy = \frac{TP + TN}{TP + TN + FP + FN}
\end{equation}

Accuracy is a poor metric for imbalanced classes because a trivial classifier can assign the majority class and achieve a high accuracy score.  To address this, the F1 score can be used instead.  For this reason, we choose to use the F1 score as our evaluation metric.  The F1 is the harmonic average of precision and recall.  The precision, recall, and F1 are computed as follows.

\begin{equation}
    Recall = \frac{TP}{TP + FN} \qquad Precision = \frac{TP}{TP + FP}
\end{equation}

\begin{equation}
    F1 = \frac{2 \times Precision \times Recall}{Precision + Recall}
\end{equation}


In a multiclass classification task, each class has its own F1 and is considered positive with the rest negative.

\section{Results}
\label{sec:results}

\subsection{ Multi-Classification }


In this section, we test the performance of models in content rating, which is a multi-classification. All the models used default hyper-parameter settings.  Two JSON files were chosen to make a comparison, where the number of a sentence per instance is 5 and 10. Each training process was repeated three times, and mean of the macro F1 score was chosen and shown in Table \ref{tab:multiclass}.

\begin{table}
\centering
\caption{Multi-class Classification Results}
\label{tab:multiclass}
\begin{tabular}{|l|l|}
\hline
\rowcolor[HTML]{EFEFEF} 
Model and Sentence Per instance & F1    \\ \hline
Naive Bayes with 5 sentences    & 0.363 \\ \hline
Naive Bayes with 10 sentences   & 0.378 \\ \hline
Random Forest with 5 sentences  & 0.382 \\ \hline
Random Forest with 10 sentences & 0.395 \\ \hline
LSTM with 5 sentences           & 0.362 \\ \hline
LSTM with 10 sentences          & 0.374 \\ \hline
ALBERT with 5 sentences         & 0.333 \\ \hline
ALBERT with 10 sentences        & 0.347 \\ \hline
DistilBERT with 5 sentences     & 0.356 \\ \hline
DistilBERT with 10 sentences    & 0.370 \\ \hline
\end{tabular}
\end{table}

All the models showed poor performance on the content rating task, among which the random forest classifier performed best. Naive Bayes, LSTM and DistilBERT owned a similar score, while ALBERT performed the worst. Meanwhile, the performance increased as the number of sentences per instance increased. However, all the F1 scores were less than 0.4, having no practical application value. Considering the time cost and effort to raise the score to a satisfactory level, I gave up the aim of content rating for fan fiction. I turned to finish a more straightforward task: identifying the fan fiction suitable for a general audience. This task transferred all categories except General Audience to Not General. In this way, it became a binary classification task. I also analysed the average word length and average sentence length in the sample set. They’re both close to a normal distribution, most of the average word length is clustered in the 5 to 6 range, and the average sentence length is usually less than 100. The distribution of sentence length varies in classes, where general audiences rating owns the shortest length, followed by teen and up to audience rating.

\subsubsection{ Mis-labelling }


While troubleshooting the poor classification results, we found many instances did not fit their rating. Though we are not professional content raters, the instance in Listing \ref{lstexample} contains an adult theme but was rated for general audiences (more blatant mis-labellings were found).  After checking 20 fiction in each category, wrong labelling frequently occurred in general audience rating: 15 of 20 could not be considered suitable for the general audience. They talked about illness, death, cherishing and dating, accompanied by slang, profanity and unsuited descriptions. The case of wrong labelling also existed in the teen, and up to audience rating, 2 of 20 should be labelled as mature rating. The inferior quality of data might be a vital factor in the poor performance of the multi-classification model. However, the guess needs the model training based on massive manual labelling samples to prove.


\subsection{ Model Tuning }

In this section, we modify certain hyper-parameters for the various models. When testing one hyper-parameter, all the others are kept unchanged. Each experiment was repeated three times, and the average value
was used for the final score.

\subsubsection{ TF-IDF Tuning }

Features for model training were built by the block of TF-IDF and chi-squared filter. The \textit{topn-features} was used to keep the top $n \in \{10000, 20000, 30000\}$ features with the highest TF-IDF scores. The chi-squared filter further reduced the features according to the \textit{p-value}. The sample set with a sentence per instance of 32 was selected because it‘s the mean of the sentence number. Feature num ranges from 10,000 to 30,000 and p-value was 0.99, 0.95 or 0.90. A naive Bayes Classifier was chosen to train the model. No consideration was given to the time cost, as the difference from changes in these parameters is small compared to the time costs of some other methods or steps.

\begin{table}
\centering
\caption{Feature Selection hyper-parameter tuning}
\label{tab:numfeatures}
\begin{tabular}{|l|l|l|l|}
\hline
\rowcolor[HTML]{EFEFEF} 
topn-features& p-value & selected & F1 \\ \hline
10000       & 0.90    & 487         & 0.653 \\ \hline
10000       & 0.95    & 187         & 0.637 \\ \hline
10000       & 0.99    & 19          & 0.609 \\ \hline
20000       & 0.90    & 716         & 0.657 \\ \hline
20000       & 0.95    & 264         & 0.651 \\ \hline
20000       & 0.99    & 26          & 0.623 \\ \hline
30000       & 0.90    & 872         & 0.659 \\ \hline
30000       & 0.95    & 327         & 0.651 \\ \hline
30000       & 0.99    & 35          & 0.632 \\ \hline
\end{tabular}
\end{table}

Table \ref{tab:numfeatures} shows the results of varying the topn-features and p-values.  Increasing topn-features slightly results in a higher F1 score. When the p-value is increased, a sharp decrease in the number of features leads to a decrease in the score. In the end, the topn-features was set to 30,000, and the p-value was set to 0.90.

\subsubsection{ Random Forest Tuning }

\begin{figure}[h]
    \centering
    \includegraphics[width=\linewidth]{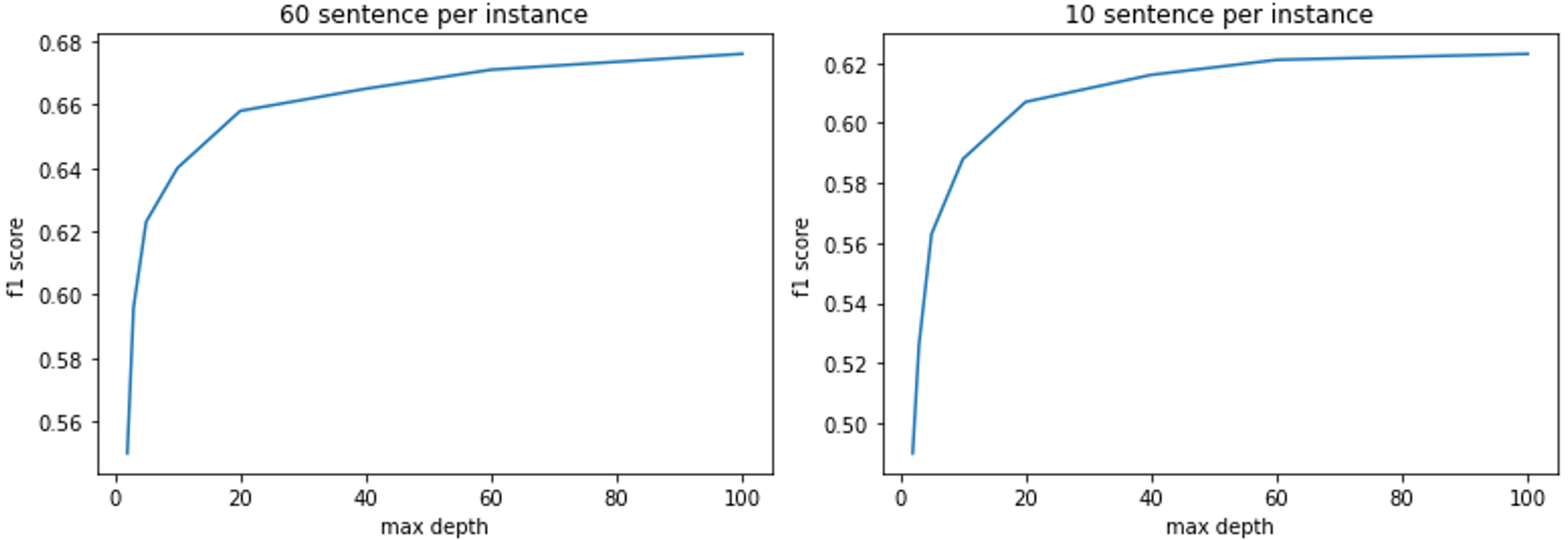}
    \caption{Random Forest Depth Tuning}
    \label{fig:tree-depth}
\end{figure}

Since the performance of the random forest model can be affected by the shape of the instance, we chose the sample sets of 10 and 60 sentences per instance to make the comparison, which reflected the situation of an extended length instance and a short length one. Figure \ref{fig:tree-depth} shows the effect that increasing the maximum tree depth has on the F1 score.  We also note, but do not show, that increasing the \textit{max\_depth} requires more computation.  There is a notable diminishing return after \textit{max\_depth} of 40 so we use this for comparison.

\begin{figure}[h]
    \centering
    \includegraphics[width=\linewidth]{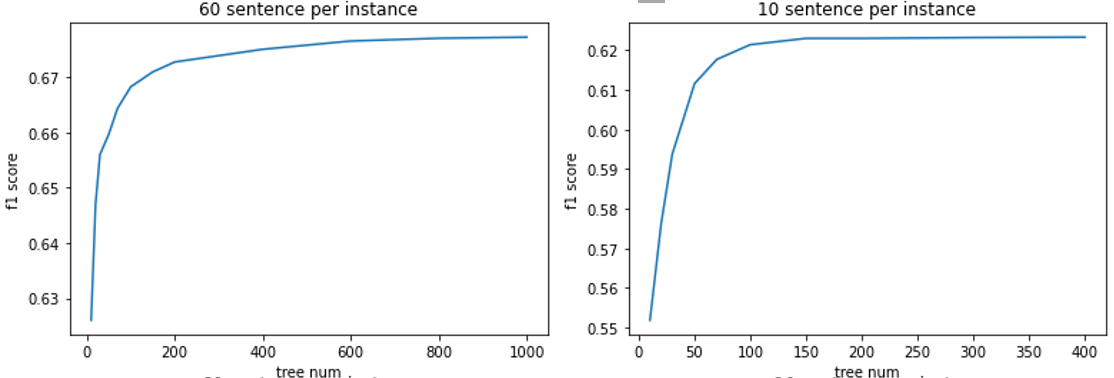}
    \caption{Random Forest Trees Tuning}
    \label{fig:tree-number}
\end{figure}

Figure \ref{fig:tree-number} similarly shows the effect of increasing the number of trees in the random forest.  For the ten and sixty-sentences per instance, the ideal value is respectively 150 and 800 trees. Meanwhile, the time cost is almost proportional to the number of trees. We set the tree number to be 20 times the number of sentences in a sample.  Unlike other hyper-parameters, the number of trees is changing with the number of samples per instance.

\subsubsection{ LSTM Tuning }

This subsection tests whether to use forward or Bi-directional LSTM (LSTM and BiLSTM respectfully) and whether to train the embedding layer.  Whether to use BiLSTM and train embedding layers were trained at first, and the epoch was adjusted to obtain a better fit. The BiLSTM was fitted to both the forward and backward input sequences, adding extra post-order information compared to LSTM. Freezing the embedding layer will stop its fine-tuning during training and save time and resource costs. As shown in Table \ref{tab:hyper_lstm}, training the embedding layer improves the score, but the decay of the score increases at more significant epochs. BiLSTM is slightly worse than LSTM in all experiments and takes longer, so LSTM was chosen.

To reduce overfitting, we tested the two methods, dropout and regularization, separately. Dropout can be described as a kind of information loss which improves the performance of a neural network by preventing feature detectors from working together.  On the other hand, regularisation exerts an additional penalty on the loss function.  Two common penalties are the L1 and the L2 (both with a $\lambda$ weighting).  We found (results omitted for brevity) the L2 penalty ($\lambda=0.01$) results in the highest F1 score of 0.651 and choose it to mitigate overfitting.  

\begin{table}
\centering
\caption{LSTM hyper-parameter tuning.Batch 128,Units 64}
\label{tab:hyper_lstm}
\begin{tabular}{|l|l|l|l|}
\hline
\rowcolor[HTML]{EFEFEF} 
LSTM\_layer & Trainable & Epoch & F1    \\ \hline
LSTM        & True      & 10    & 0.598 \\ \hline
BILSTM      & True      & 10    & 0.596 \\ \hline
LSTM        & True      & 3     & 0.648 \\ \hline
BILSTM      & True      & 3     & 0.644 \\ \hline
LSTM        & False     & 10    & 0.604 \\ \hline
BILSTM      & False     & 10    & 0.596 \\ \hline
LSTM        & False     & 3     & 0.633 \\ \hline
BILSTM      & False     & 3     & 0.627 \\ \hline
\end{tabular}
\end{table}

\subsubsection{ BERT-DLM Tuning }

\begin{table}[]
\centering
\caption{BERT-DLM hyper-parameter tuning.}
\label{tab:hyper_bert}
\begin{tabular}{|l|l|l|l|}
\hline
\rowcolor[HTML]{EFEFEF} 
Batch size & Dense unit & Time & Score \\ \hline
8          & 64         & 1062 & 0.613 \\ \hline
16         & 64         & 516  & 0.614 \\ \hline
32         & 64         & 304  & 0.609 \\ \hline
64         & 64         & 162  & 0.596 \\ \hline
16         & 16         & 442  & 0.611 \\ \hline
16         & 32         & 486  & 0.613 \\ \hline
\end{tabular}
\end{table}

We consider the batch and number of units in the dense layer for the BERT-DLM model.  Table \ref{tab:bert-dlm} shows the effect that varying these hyper-parameters has on the F1 score.  Based on this we choose dense unit=64 and batch size=16. We found the BERT-DLM model more unstable than the other models.

\begin{figure}[h]
    \centering
    \includegraphics[width=0.62\linewidth]{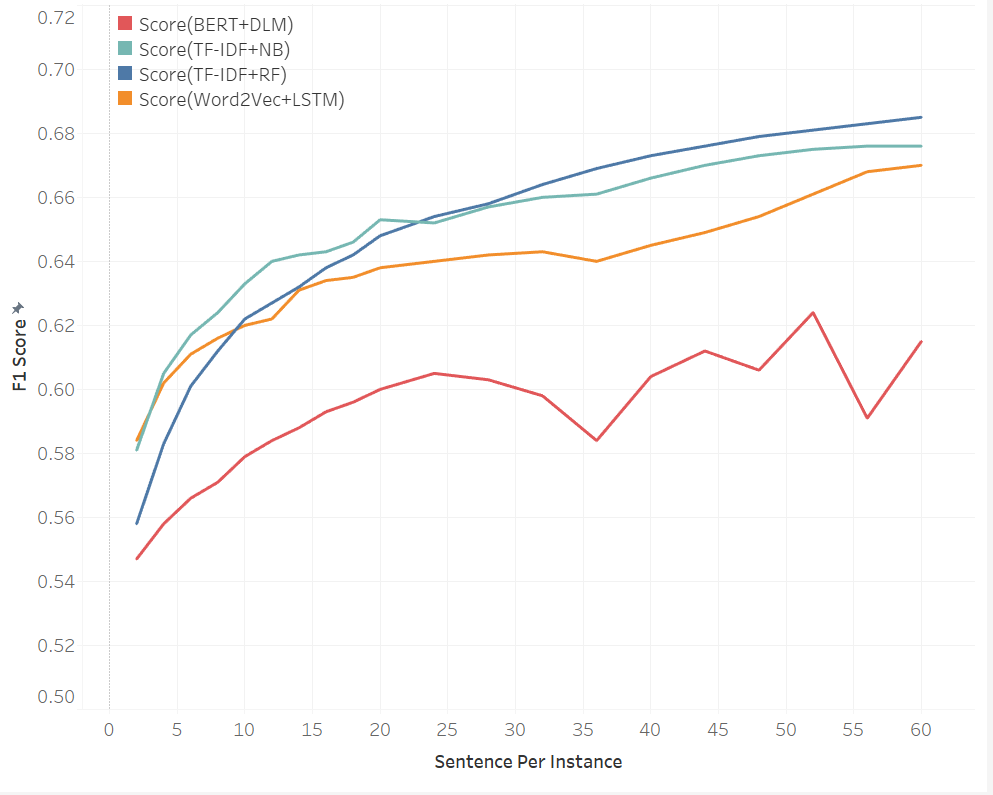}
    \caption{Model Accuracy Comparison}
    \label{fig:model_comparison}
\end{figure}

\subsection{ Model Comparison }

\begin{figure}[h]
    \centering
    \includegraphics[width=0.62\linewidth]{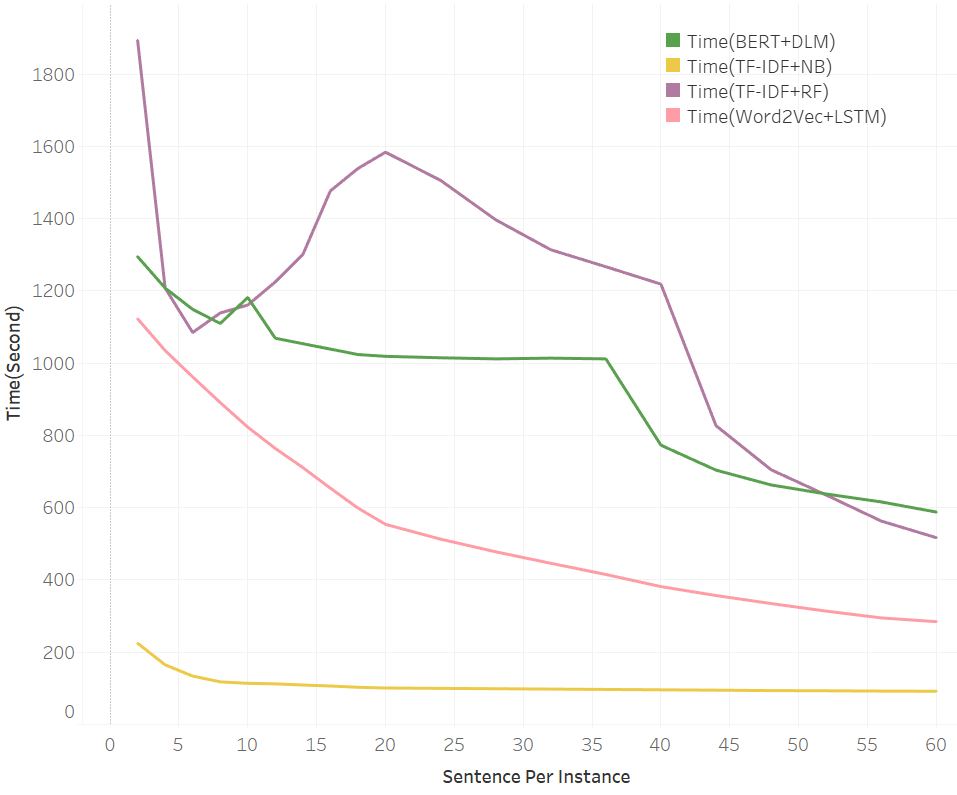}
    \caption{Model Computation Comparison}
    \label{fig:model_times}
\end{figure}

In this section, we train all the datasets obtained by segmenting them according to the number of sentences and compare and analyse the results. The experiment was repeated three times and averaged. As shown in the figure, as the number of sentences per instance increases, all methods except the BERT method achieve a smooth growth, which is also true at first but starts to fluctuate in the score when the number of sentences exceeds 20. Its growth dramatically improves performance when the sentence per instance is small. This improvement gradually slows down, with the best-performing TF-IDF+RF method, for example, bringing a score improvement of about 9\% as the number of sentences grows from 2 to 20. However, a score improvement of less than 5\% from 20 to 60, achieving a best F1 score of 0.685. Results for different methods illustrate the strengths and weaknesses of the various feature extraction methods in handling this task, where TF-IDF performs best. The results of the TF-IDF+RF method keep increasing till the end, which means higher sentences may bring better performance.


\figurename~\ref{fig:model_times} shows the time consumed for each model to include feature extraction,  model training, and model prediction. The same computer is used for the experiments. According to the experiment results, the TF-IDF+NB method maintains the shortest elapsed time throughout the process, except for the three methods, which all have a long-elapsed time when the sentences per instance is small. As it increases, the time decreases rapidly, but the TF-IDF+RF method shows some fluctuations because the number of trees hyper-parameter is a function of the number of sentences, unlike the other hyper-parameters that remain fixed. At sentences=60, the best results are obtained for all methods except BERT, where the most time-consuming method, BERT+DLM, takes about six times longer than TF-IDF+NB, and the best-performing method, TF-IDF+RF, takes five times longer.

\section{Conclusion}
\label{sec:conclusion}

In this paper considered an approach to classifying the content rating of fan fiction. We created and implemented a web crawler to obtain the fan fiction along with the self-annotations.  We described the data pre-processing and two traditional and two deep learning architectures. We found the  multi-classification accuracy to be quite poor so decided to use a binary classifier to identify general audience versus not general audience.  To properly compare the architectures, we performed hyper-parameter tuning of each architecture.  We then show that the traditional approaches were quite competitive in terms of accuracy and computational performance.  Finally, we submit that incorrect labelling will need to be addressed for future work to be able to significantly improve accuracy performance.


\bibliographystyle{unsrt}  
\bibliography{references}

\end{document}